\documentclass[conference]{IEEEtran}
\IEEEoverridecommandlockouts

\usepackage{cite}
\usepackage{float}
\usepackage{amsmath,amssymb,amsfonts}
\usepackage{algorithmic}
\usepackage{graphicx}
\usepackage{textcomp}
\usepackage{xcolor}
\usepackage{booktabs}
\usepackage{multirow}
\usepackage[hidelinks]{hyperref}
\usepackage{svg}

\DeclareMathOperator*{\argmin}{argmin}
\def\BibTeX{{\rm B\kern-.05em{\sc i\kern-.025em b}\kern-.08em
    T\kern-.1667em\lower.7ex\hbox{E}\kern-.125emX}}
\begin{document}

\title{Complete Suffix Prediction for Recommendation via Latent Retrieval over Process Graphs\\
}

\author{\IEEEauthorblockN{Sarra MADAD}
\IEEEauthorblockA{\textit{LIST3N Lab, AI Lab} \\
\textit{University of Technology of Troyes (UTT), QAD}\\
Paris, France \\
sarra.madad@qad.com}
\\
\IEEEauthorblockN{Frédéric BERTRAND}
\IEEEauthorblockA{\textit{CEDRIC} \\
\textit{Conservatoire national des arts et métiers (Cnam)}\\
Paris, France \\
frederic.bertrand@lecnam.net}
\and
\IEEEauthorblockN{Myriam MAUMY-BERTRAND}
\IEEEauthorblockA{\textit{Arènes (UMR CNRS 605)} \\
\textit{Ecole des hautes études en santé publique (EHESP)}\\
Rennes, France \\
myriam.maumy@ehesp.fr}
\\
\IEEEauthorblockN{Yoann VALERO}
\IEEEauthorblockA{\textit{AI Lab} \\
\textit{QAD}\\
Strasbourg, France \\
yoann.valero@qad.com}
}

\maketitle

\begin{abstract}
Complete suffix prediction is challenging in sequential decision settings, where the same prefix can remain compatible with several plausible suffixes. We propose a graph-based metric-learning framework that reformulates complete suffix prediction as latent retrieval over process graphs. Prefixes and suffixes are represented as directed attributed graphs and encoded by edge-conditioned graph neural networks, allowing event-level activities and transition-level durations to be modelled jointly. Prefix representations are projected into the latent suffix space through a predictor trained with a joint reconstruction and contrastive objective strengthened using process-aware hard negatives. To stabilise the learned retrieval geometry, spectral normalisation, and retrieval robustness, spectral normalisation is applied to enforce a Lipschitz constraint on both encoders and predictor. 
Experiments on two real-life process datasets demonstrate that the proposed framework achieves the best overall results across nearly all evaluated criteria. It improves semantic suffix accuracy measured by normalized Damerau-Levenshtein distance, yields strong retrieval quality through Recall@1, Recall@5, and MRR@5, and maintains temporal plausibility according to Mean Absolute Error. These results show that graph-based latent retrieval is an effective alternative to sequential suffix prediction for recommendation-oriented process monitoring under structural and KPI-related constraints.


\end{abstract}


\section{Introduction}

Predicting long-term  process trajectories under high uncertainty is a central challenge in prescriptive process monitoring \cite{vanderAalst2022, DBLP:journals/corr/abs-2112-01769, articleMarquez}. Given an ongoing case prefix, the goal is not only to predict the next event, but to infer plausible complete future continuations that can support KPI-oriented decisions \cite{articleTeinemaa}.

Predictive process monitoring has evolved from classical machine learning models based on hand-crafted prefix features \cite{DBLP:journals/corr/abs-1905-09568, articleFrancescomarino, inproceedingsTerragni} to sequential neural architectures such as RNNs, LSTMs, and transformers \cite{articleHochreiter, Tax_2017, vaswani2023attentionneed, DBLP:journals/corr/abs-2104-00721, inproceedingsWuyts}. While these models improved next-activity and remaining-time prediction, prescriptive monitoring requires reasoning over complete future trajectories rather than local predictions \cite{taymouri2021deepadversarialmodelsuffix, articleGunnarsson}.

Existing complete suffix prediction methods suffer from two limitations. First, most approaches output a single deterministic future, either through autoregressive decoding or direct suffix prediction. This creates a single-output bottleneck: in stochastic processes, several distinct continuations may remain plausible from the same prefix \cite{inproceedingsKunkler}. Second, they must jointly optimise structural accuracy and temporal prediction, typically through parallel losses for activities and timestamps \cite{Tax_2017, inproceedingsWuyts}. These objectives only partially align leading to interference between semantic fidelity and temporal plausibility.

We address these limitations by reformulating complete suffix prediction as a graph-based metric-learning problem. Instead of decoding a single suffix, the proposed model learns a retrieval geometry in which a prefix embedding is projected close to several plausible complete suffixes. Prefixes and historical suffixes are encoded as directed attributed process graphs through edge-conditioned GNNs \cite{10.1007/978-3-030-94343-1_3}, so that event-level activity information and transition-level temporal information are represented jointly. The prefix embedding is then mapped into the suffix space and used as a query for nearest-neighbour KPI-aware retrieval \cite{DBLP:journals/corr/abs-2004-04523}. Training combines latent reconstruction with a process-aware contrastive objective using hard negatives selected among behaviourally plausible suffixes \cite{NEURIPS2020_d89a66c7}. To stabilise this retrieval geometry and reduce distortions between structural and temporal signals, we impose a Lipschitz constraint through spectral normalisation \cite{gouk2020regularisationneuralnetworksenforcing}.

Experiments on a large-scale process dataset, with up to 688\,396 candidate suffixes, show that the proposed framework retrieves accurate and actionable futures. It achieves a top-1 Damerau–Levenshtein distance of 0.0976, a Recall@1 of 0.5654 on complex minority trajectories, and a temporal MAE only 8 hours above the theoretical stochastic floor under a strict joint semantic-temporal criterion. The main contributions are:

1. To formulate complete suffix prediction as graph-based latent retrieval, replacing single-suffix decoding with top-\(k\) retrieval of plausible complete continuations.

2. To introduce a process-aware metric-learning objective that combines latent reconstruction with contrastive hard negatives selected from behaviourally similar but distinct suffixes.

3. To integrate edge-conditioned process graph encoders with Lipschitz-oriented spectral normalisation, stabilising the suffix retrieval space under structural and temporal variations.

4. To evaluate the resulting retrieval space through ranking, semantic, temporal, ablation, and KPI-aware recommendation criteria on large real-world process logs.

\section{Related Work}


\subsection{Predictive Signals for Prescriptive Process Monitoring}

Prescriptive process monitoring aims to support decisions during process execution \cite{DBLP:journals/corr/abs-2112-01769}. Most approaches rely on predictive signals extracted from an ongoing prefix, such as next activity, remaining time \cite{Tax_2017, inproceedingsWuyts}, deviation risk, or outcome probability \cite{articleMarquez}. Early work used rule-based systems \cite{DBLP:journals/corr/abs-1905-09568}, similarity-based methods \cite{articleFrancescomarino, Weinzierl_2020}, and supervised learning on hand-crafted prefix features \cite{articleKim, articleMarquez}. These methods are interpretable but limited in their ability to capture temporal dependencies and process variability.

Sequential neural models, including RNNs, LSTMs, and transformers \cite{articleHochreiter, Tax_2017, vaswani2023attentionneed, DBLP:journals/corr/abs-2104-00721, inproceedingsWuyts}, improved the modelling of execution dynamics. However, the information they provide often remains local or aggregated. In prescriptive settings, next-activity, remaining-time or outcome prediction is often insufficient: decision support requires access to complete possible continuations that can be compared and ranked.

\subsection{Complete Suffix Prediction}

Complete suffix prediction addresses this limitation by predicting the full continuation of a case from its prefix. Existing methods can be grouped into two families. The first relies on sequential forecasting, using RNNs, LSTMs, GRUs, seq2seq models and transformer decoders \cite{articleHochreiter, NIPS2014_5a18e133, taymouri2021deepadversarialmodelsuffix}. These models generate suffixes step by step, but remain exposed to autoregressive error accumulation.

The second family predicts suffixes more directly, without explicit token-by-token generation \cite{articleGunnarsson, inproceedingsWuyts}. Such approaches better reflect the structured nature of complete suffix prediction and reduce local error propagation. However, both families usually return a single predicted continuation. In stochastic processes, several futures may remain plausible from the same prefix \cite{inproceedingsKunkler}; returning only one suffix hides relevant alternatives and limits recommendation.

\subsection{Joint Sequence and Time Prediction}

Temporal prediction, such as remaining time or total duration, has been widely studied in process mining \cite{DBLP:journals/corr/abs-2112-01769, articleMarquez}. Deep learning models later introduced joint prediction of activities and timestamps through multi-task learning \cite{lecun2015deep, Tax_2017}, and transformer-based models extended this idea with attention mechanisms \cite{articleGunnarsson, inproceedingsWuyts, DBLP:journals/corr/abs-2104-00721}.

Complete suffix prediction requires jointly modelling a discrete activity sequence and continuous temporal quantities. Existing architectures typically optimise parallel losses, such as cross-entropy for activities and MAE-based losses for timestamps or remaining time \cite{Tax_2017, taymouri2021deepadversarialmodelsuffix, articleGunnarsson, inproceedingsWuyts, Weinzierl_2020, agarwal2022goalorientedbestactivityrecommendation}. These objectives are only partially aligned: improving temporal fit does not necessarily improve structural correctness. This creates multi-objective interference, especially over long horizons, and motivates alternatives to single sequential decoding.

\subsection{Latent Retrieval and Metric Learning}

Retrieval-based approaches provide an alternative to single-output prediction \cite{inproceedingsKunkler}. Instead of generating one continuation, they learn an embedding space in which a prefix is mapped close to relevant future candidates. This family is related to metric learning, siamese architectures, triplet loss and contrastive learning with hard negative mining \cite{NEURIPS2020_d89a66c7, DBLP:journals/corr/SchroffKP15, 1467314}. Its main advantage is that several plausible futures can coexist in the neighbourhood of the same query.

Process mining already includes related ideas through similarity-based recommendation, nearest-neighbour retrieval and case-based reasoning over historical traces \cite{articleFrancescomarino, Weinzierl_2020, inproceedingsKunkler}. However, these approaches often rely on manually designed similarity functions and remain under-explored for scalable complete suffix retrieval under both semantic and temporal constraints.
Our work differs from standard metric-learning and retrieval-based recommendation in three respects. 

First, the objects to be retrieved are not independent items but complete process continuations with ordered activities and transition durations. As a result, we avoid autoregressive error propagation, mitigate the single-output bottleneck, and preserve multiple plausible future candidates. 

Second, similarity is not learned from generic labels only: negative suffixes are selected using process-aware distances that account for duration, length, activity presence, transition timing, and ordering. 

Third, prefixes and suffixes are encoded as attributed process graphs, allowing transition attributes to modulate message passing rather than being appended only as sequential covariates. The resulting framework, therefore, combines graph representation learning, metric learning, and retrieval-based recommendation in a suffix-prediction setting where the retrieved candidates must remain both semantically valid and temporally plausible and support KPI-oriented prescriptive recommendation.


\section{Methodology}

\subsection{Framework Overview}

The proposed framework learns a latent space for retrieving complete process suffixes. Given a prefix, its graph representation \cite{skenderi2025graphlevelrepresentationlearningjointembedding} is encoded by a graph neural network \cite{10.1007/978-3-030-94343-1_3} and projected through an MLP into the suffix latent space \cite{DBLP:journals/corr/SchroffKP15, 1467314}. This embedding is used as a query to retrieve the nearest encoded suffix candidates.

Training combines a reconstruction-oriented objective with a contrastive loss using process-aware negative suffixes \cite{NEURIPS2020_d89a66c7}. Lipschitz regularisation \cite{gouk2020regularisationneuralnetworksenforcing} stabilises the latent geometry, supporting robust, temporally plausible suffix retrieval for KPI-aware recommendation \cite{articleTeinemaa}.

\subsection{Problem Formulation}

Let \(\mathcal E\) denote the event space, and let \(\mathcal E^*\) denote the Kleene closure of \(\mathcal E\), i.e., the set of all finite event sequences over \(\mathcal E\). An event log is defined as a collection of traces $\smash{\mathcal L = \{\sigma^{(1)}, \sigma^{(2)}, \dots, \sigma^{(N)}\} \subseteq \mathcal E^*}$ with \(\smash{N = |\mathcal L|}\), where each trace \(\smash{\sigma^{(c)}}\) is a finite ordered sequence of events $\smash{\sigma^{(c)} = (e_{1}^{(c)}, e_{2}^{(c)}, \dots, e_{n_c}^{(c)})}$ with \(\smash{e_t^{(c)} \in \mathcal E}\) for all \(\smash{t \in \{1,\dots,n_c\}}\), and \(\smash{n_c = |\sigma^{(c)}|}\). Each event \(\smash{e_t^{(c)}}\) is associated with an activity label, a timestamp, and optional contextual attributes. We assume that traces are fully observed and temporally ordered, i.e., they do not contain missing events or internal gaps. More precisely, letting \(\smash{\tau: \mathcal E \to \mathbb{R}}\) denote the timestamp projection, the ordering of events within a trace satisfies $\smash{\tau(e_r^{(c)}) < \tau(e_s^{(c)})}$ whenever $r < s$.

For a trace $\smash{\sigma^{(c)} = (e_{1}^{(c)}, e_{2}^{(c)}, \dots, e_{n_c}^{(c)})}$, we define a prefix of length $\smash{m\in \{1,\dots,{n_c}\}}$ as $\smash{\pi^{(c)}_{(m)} = (e_{1}^{(c)}, e_{2}^{(c)},\dots,}$ $\smash{e_{m}^{(c)}) \in \mathcal{E}^*}$ and its corresponding suffix as $\smash{\gamma^{(c)}_{(m)}} = \smash{(e^{(c)}_{m+1}, e^{(c)}_{m+2}, \dots, e_{n_c}^{(c)}) \in \mathcal{E}^*}$. Hence, $|\smash{\pi^{(c)}_{(m)}}| = m$ and $|\gamma^{(c)}_{(m)}| = n_c-m$.
Given a prefix \(\smash{\pi^{(c)}_{(m)}}\), the complete suffix prediction task consists in identifying a plausible future continuation \(\smash{\hat{\gamma}^{(c)}_{(m)}}\) among a set of candidate suffixes. In contrast with autoregressive forecasting, our objective is not to generate the suffix event by event, but to retrieve a complete suffix from a structured latent space.

We associate each prefix and suffix with graph representations
$\smash{G^{p}_{(c)}} = \smash{\phi_p(\pi^{(c)}_{(m)})} \smash{\in \mathcal{G}}$, $\smash{G^{s}_{(c)}} = \smash{\phi_s(\gamma^{(c)}_{(m)})} \smash{\in \mathcal{G}}$,
where \(\smash{\mathcal{G}}\) denotes the space of directed attributed multigraphs, and $\smash{\phi_p : \mathcal{E}^* \to \mathcal{G}}$,
$\smash{\phi_s : \mathcal{E}^* \to \mathcal{G}}$ map event sequences to their corresponding prefix and suffix directed multigraphs.

Let $\smash{f_p:\mathcal{G}\to\mathbb{R}^{d_{\mathrm{z}}}}$ and $\smash{f_s:\mathcal{G}\to\mathbb{R}^{d_{\mathrm{z}}}}$ denote the prefix and suffix encoders, respectively. We then define the latent embeddings of the prefix and suffix graphs produced by their respective encoders:
$\smash{z^{(c)}_p = f_p(G^{p}_{(c)}) \in \mathbb{R}^{d_{\mathrm{z}}}}$ and 
$\smash{z^{(c)}_s = f_s(G^{s}_{(c)}) \in \mathbb{R}^{d_{\mathrm{z}}}}$
where \(\smash{d_{\mathrm{z}}}\) denotes the dimension of the shared latent space, chosen to provide a compact representation while preserving sufficient discriminative capacity to separate structurally distinct process graphs.

A predictor $\smash{h : \mathbb{R}^{d_z} \to \mathbb{R}^{d_z}}$ maps the prefix embedding into the suffix latent space $\smash{\hat{z}^{(c)}_s = h(z^{(c)}_p)}$ where \(\smash{\hat{z}^{(c)}_s}\) is the predicted latent representation of the future suffix. It acts as a query point in the latent suffix space.

At inference time, complete suffix prediction is formulated as a nearest-neighbour retrieval problem in this space. Let
$\mathcal{C}=\left\{\gamma_1,\ldots,\gamma_K\right\}$, 
$K = |\mathcal{C}|$, and 
$\mathcal{C} \subseteq \mathcal{E}^*$ 
be the candidate set of suffixes, where each candidate suffix \(\gamma_j\) is associated with a latent embedding
$z_{s_j}\in\mathbb{R}^{d_z}$.
Let $d:\mathbb{R}^{d_z}\times\mathbb{R}^{d_z}\to\mathbb{R}_{+}$ denote the latent distance. Given the query embedding \(\smash{\hat{z}^{(c)}_s}\), the model retrieves the \(k\) nearest candidate suffixes:
$\smash{\mathcal{N}_k(\hat{z}^{(c)}_s)
=
\operatorname{TopK}_{\gamma_j \in \mathcal{C}}
\left(
-d(\hat{z}^{(c)}_s, z_{s_j})
\right)}$ with $\smash{\mathcal{N}_k(\hat{z}^{(c)}_s)\subseteq \mathcal{C}.}
\label{eq:topk_retrieval}$

The top retrieved candidate is obtained as the first element of \(\smash{\mathcal{N}_k(\hat{z}^{(c)}_s)}\), or equivalently
$\smash{\hat{\gamma}^{(c)}_{(m)}
=
\smash{\argmin_{\gamma_j \in \mathcal{C}}
d(\hat{z}^{(c)}_s, z_{s_j})
\label{eq:top1_retrieval}.}}$ 
Beyond top-1 retrieval, the set \(\mathcal{N}_k(\hat{z}^{(c)}_s)\) can also be used for recommendation. Let $q:\mathcal{C}\to\mathbb{R}$ denote a target KPI scoring function. A KPI-aware recommendation is then obtained by selecting, among the retrieved plausible suffixes, the candidate that optimises the target KPI, for instance the total trace duration:
$\smash{{\gamma^{\star}_{\mathrm{KPI}}
=
\argmin_{\gamma_j \in \mathcal{N}_k(\hat{z^{(c)}_s})} q(\gamma_j)
\label{eq:kpi_retrieval}.}}$
\\

In the absence of a KPI-aware selection step, the final predicted suffix is given by the nearest neighbour in~\eqref{eq:top1_retrieval}. When recommendation is KPI-aware, the final selected suffix is instead chosen among the top-\(k\) retrieved candidates according to the objective \(q(\gamma_j)\), as defined in~\eqref{eq:kpi_retrieval}.
The retrieved suffixes are further evaluated along two complementary axes:
    \textbf{i)} \textbf{semantic accuracy} \cite{zhao2019string}, measured through the normalised Damerau-Levenshtein distance (N DLD) between the retrieved suffix and the ground-truth one;
    \textbf{ii)} \textbf{temporal plausibility} \cite{articleVerenich}, measured through the absolute error between the real suffix duration and the duration associated with the retrieved suffix.

For clarity, we use fully indexed notations in the present subsection to specify the task at the trace level. In the rest of the paper, whenever the considered trace and prefix length are clear from context, we adopt lighter notations and omit some superscripts or subscripts to avoid unnecessary clutter.

\subsection{Data Construction and Integrity}

This section describes how the event log, the prefix--suffix pairs, and their graph representations introduced above are constructed in practice, while enforcing integrity constraints designed to prevent information leakage and preserve distributional consistency across the train, validation, and test splits.

\subsubsection*{\textbf{Event Log Cleaning and Compliance Filtering}}

Starting from the raw event log, we retain only complete and compliant traces according to domain-specific process rules. These rules enforce consistency with the expected process semantics, including constraints on terminal activities, item categories, and key ordering relations between events. After compliance filtering, we remove trace lengths with insufficient statistical support, thereby avoiding extremely sparse trajectory patterns that would otherwise destabilise suffix learning and retrieval.
Finally, an explicit terminal state \texttt{END} is appended to each trace in order to make process completion observable in the learned suffix space.

\subsubsection*{\textbf{Temporal Enrichment of Events}}

Each event is augmented with a normalised inter-event duration and cyclical temporal encodings (hour, weekday, and day of year) using sine/cosine transformations. These continuous features preserve periodicity and are well suited for neural encoders \cite{khazem2025cyclicaltemporalencodinghybrid}.

\subsubsection*{\textbf{Prefix--Suffix Construction and Graph Conversion}}
For each trace, we generate all valid prefix--suffix pairs \((\pi^{(m)},\gamma^{(m)})\) satisfying minimum length constraints and indexed by case identifier and prefix length. Each prefix and suffix is then mapped to a directed attributed multigraph, \(G^p=\phi_p(\pi^{(m)})\) and \(G^s=\phi_s(\gamma^{(m)})\), where nodes represent events and edges connect temporally consecutive events. Node attributes combine activity labels with cyclical temporal descriptors, while edge attributes encode inter-event durations. 
Representing prefixes and suffixes as attributed graphs therefore enables separating explicitly what is attached to events from what is attached to transitions while preserving the ordered relational structure of the process trajectory.
The resulting graphs are converted into \texttt{PyTorch Geometric} data objects and used as inputs to the encoders.

\subsubsection*{\textbf{Split Integrity and Distributional Control}}

To prevent information leakage, training, validation and test splits are built at the \emph{case level}: all prefix--suffix pairs from the same trace are assigned to the same split \cite{DBLP:journals/corr/abs-2107-01905}. Although chronological splitting is preferable for strict online forecasting, our goal is to learn a coherent latent space of complete future continuations.
We also enforce integrity constraints on the training split. It must cover the full activity vocabulary and preserve sufficient support across trace lengths and non-singleton behavioural variants. Representativeness is monitored through Jensen--Shannon divergence between the training split and the full dataset over activity and variant distributions \cite{SALAZAR2022109885}. The split may be corrected until set divergence thresholds are satisfied.

\subsubsection*{\textbf{Duration Normalisation and Contrastive Negative Construction}}

Durations, used as edge attributes, are normalised using training statistics only: values are clamped to be non-negative, transformed with \(\smash{\log(1+x)}\), and standardised. Cyclical temporal node attributes are encoded with sine and cosine functions and are therefore bounded in \(\smash{[-1,1]}\).

For each positive suffix graph \(\smash{G^s}\), we compute a bounded process-aware distance to training candidate suffixes that combines normalised discrepancies in total duration, suffix length, activity presence, transition durations, and ordering. Self-matches and suffixes from the same case are excluded.

Negatives are not sampled uniformly. Candidate suffixes are ranked by process-aware distance, and the negative is sampled from a predefined rank band within the top-\(K\) pool. This yields negatives that are neither trivial nor overly deterministic, while controlling their difficulty. A quota limits how often each suffix can be reused as a negative. The resulting triplets \(\smash{(G^p, G^s, G^-)}\) are used to train the latent retrieval model.

Rarity-aware weights are computed from training frequencies to rebalance under-represented activities during learning.

\subsection{Prefix and Suffix Encoders}
Each prefix and suffix is represented as a directed attributed multigraph \cite{skenderi2025graphlevelrepresentationlearningjointembedding}, denoted \(\smash{G=(V,E,X_V^{\mathrm{raw}},X_E)}\), with \(\smash{X_V^{\mathrm{raw}}\in\mathbb{R}^{|V|\times d_v^{\mathrm{raw}}}}\) and \(\smash{X_E\in\mathbb{R}^{|E|\times d_e}}\). Nodes correspond to events and edges to temporally consecutive transitions. For a node \(\smash{v}\), the raw feature vector \(\smash{x_v^{\mathrm{raw}}\in\mathbb{R}^{d_v^{\mathrm{raw}}}}\) combines activity identity and contextual temporal descriptors, so that \(\smash{d_v^{\mathrm{raw}}=1+d_t}\). For an edge \(\smash{(v_i,v_j)}\), the feature vector \(\smash{x_{v_iv_j}\in\mathbb{R}^{d_e}}\) encodes transition-level information, which in our case is the normalised elapsed time between the source node \(\smash{v_i}\) and the destination node \(\smash{v_j}\).

Let \(\smash{\mathcal{A}}\) denote the activity alphabet. Each activity label \(\smash{a_v\in\mathcal{A}}\) is mapped to a learnable embedding \(\smash{\mathrm{Emb}:\mathcal{A}\to\mathbb{R}^{\lceil\sqrt{|\mathcal{A}|}\rceil}}\). The initial node representation is defined as \(\smash{h_v^{0}=\mathrm{LN}([\mathrm{Emb}(a_v)\,\|\,t_v])\in\mathbb{R}^{d_v}}\), where \(\smash{t_v\in\mathbb{R}^{d_t}}\) contains the cyclical encodings of hour-of-day, day-of-week, and day-of-year, and \(\smash{d_v=\lceil\sqrt{|\mathcal{A}|}\rceil+d_t}\). By stacking all node representations, we obtain \(\smash{H_V^{(0)}\in\mathbb{R}^{|V|\times d_v}}\).

Two graph encoders are used, \(\smash{z_p=f_p(G^p)\in\mathbb{R}^{d_z}}\) and \(\smash{z_s=f_s(G^s)\in\mathbb{R}^{d_z}}\), where \(\smash{f_p}\) and \(\smash{f_s}\) are two-layer edge-conditioned message-passing networks \cite{simonovsky2017dynamicedgeconditionedfiltersconvolutional}. This choice is motivated by the fact that process continuations are structured objects in which both event-level and transition-level information matter. In particular, transition attributes such as inter-event durations should directly modulate message propagation, which is not captured by fixed propagation rules as in standard GCN layers \cite{DBLP:journals/corr/KipfW16}.

At each layer \(\smash{\ell\in\{1,2\}}\), the encoder applies an edge-conditioned message function \(\smash{\phi^{(\ell)}}\), an aggregation operator, and an update function \(\smash{\psi^{(\ell)}}\):
\begin{equation}
\begin{aligned}
h_{v_j}^{(\ell)}
&=
\psi^{(\ell)}\big(
h_{v_j}^{(\ell-1)},
\operatorname{AG}_{v_i \in \mathcal{N}(v_j)},
\phi^{(\ell)} (
h_{v_j}^{(\ell-1)},\,
h_{v_i}^{(\ell-1)},\,
x_{v_i v_j}
)
\big)
\end{aligned}
\label{eq:mpnn_general}
\end{equation}

In our implementation, this scheme is instantiated with {\tt NNConv} layers. The edge attribute \(\smash{x_{uv}}\) is processed by a small edge network \(\smash{g^{(\ell)}}\), defined as:
\begin{equation}
g^{(\ell)}(x)=\mathrm{SN}(L_2^{(\ell)})(\mathrm{LeakyReLU}(\mathrm{SN}(L_1^{(\ell)})(x)))    
\end{equation}
where \(\smash{L_1^{(\ell)}}\) and \(\smash{L_2^{(\ell)}}\) are the two linear layers of the edge network and \(\mathrm{SN}(\cdot)\) denotes spectral normalisation. The first layer maps the edge attribute to a hidden edge representation, while the second outputs a flattened filter reshaped into the edge-conditioned matrix \(\smash{W_{v_iv_j}^{(\ell)}}\). In particular, \(\smash{W_{v_iv_j}^{(1)}\in\mathbb{R}^{d_h\times d_v}}\) and \(\smash{W_{v_iv_j}^{(2)}\in\mathbb{R}^{d_h\times d_h}}\), where \(\smash{d_h}\) is the hidden node dimension. In our implementation, the graph-level latent dimension is set equal to the hidden node dimension, i.e. \(\smash{d_z=d_h}\).

Messages are aggregated by mean pooling as \(\smash{m_{v_j}^{(\ell)}}=\smash{\frac{1}{|\mathcal{N}(v_j)|}\sum_{v_i\in\mathcal{N}(v_j)}W_{v_iv_j}^{(\ell)}h_{v_i}^{(\ell-1)}}\), then combined with a root transformation \(\smash{U^{(\ell)}h_{v_j}^{(\ell-1)}}\) and passed through a LeakyReLU activation. The first layer maps node features from \(\smash{d_v}\) to \(\smash{d_h}\), while the second preserves the hidden dimension. Spectral normalisation controls the sensitivity of the edge network to duration perturbations, and LeakyReLU preserves this Lipschitz control while introducing non-linearity.
After the second layer, node embeddings are aggregated through global mean pooling, \(\smash{z=\frac{1}{|V|}\sum_{v_j\in V}h_{v_j}^{(2)}\in\mathbb{R}^{d_z}}\). This vector is used as the graph-level latent representation of the prefix or suffix. Using two distinct encoders allows the model to specialise the representation of observed prefixes and future suffixes while still projecting them into a comparable latent space.

\subsection{Prefix-to-Suffix Latent Prediction}
Given a prefix embedding $\smash{z_p}$, a predictor $h$ maps it into the suffix latent space $\smash{\hat{z}_s = h(z_p)}$.
The predictor is implemented as a two-layer multilayer perceptron with spectral normalisation and ReLU activation: 
$h(z_p)=\mathrm{SN}(W_2)\,\rho\!\left(\mathrm{SN}(W_1)z_p+b_1\right)+b_2$ 
where \(\smash{\rho(\cdot)=\mathrm{ReLU}(\cdot)}\), \(\smash{W_1,W_2\in\mathbb{R}^{d_z\times d_z}}\), and \(\smash{b_1,b_2\in\mathbb{R}^{d_z}}\) denote the bias vectors of the two linear layers. Since the graph-level latent dimension coincides with the hidden dimension, the predictor preserves dimensionality, i.e., \(\smash{d_z=d_h}\).

\begin{figure*}[!ht]
  \centering
  \includegraphics[scale=0.16]{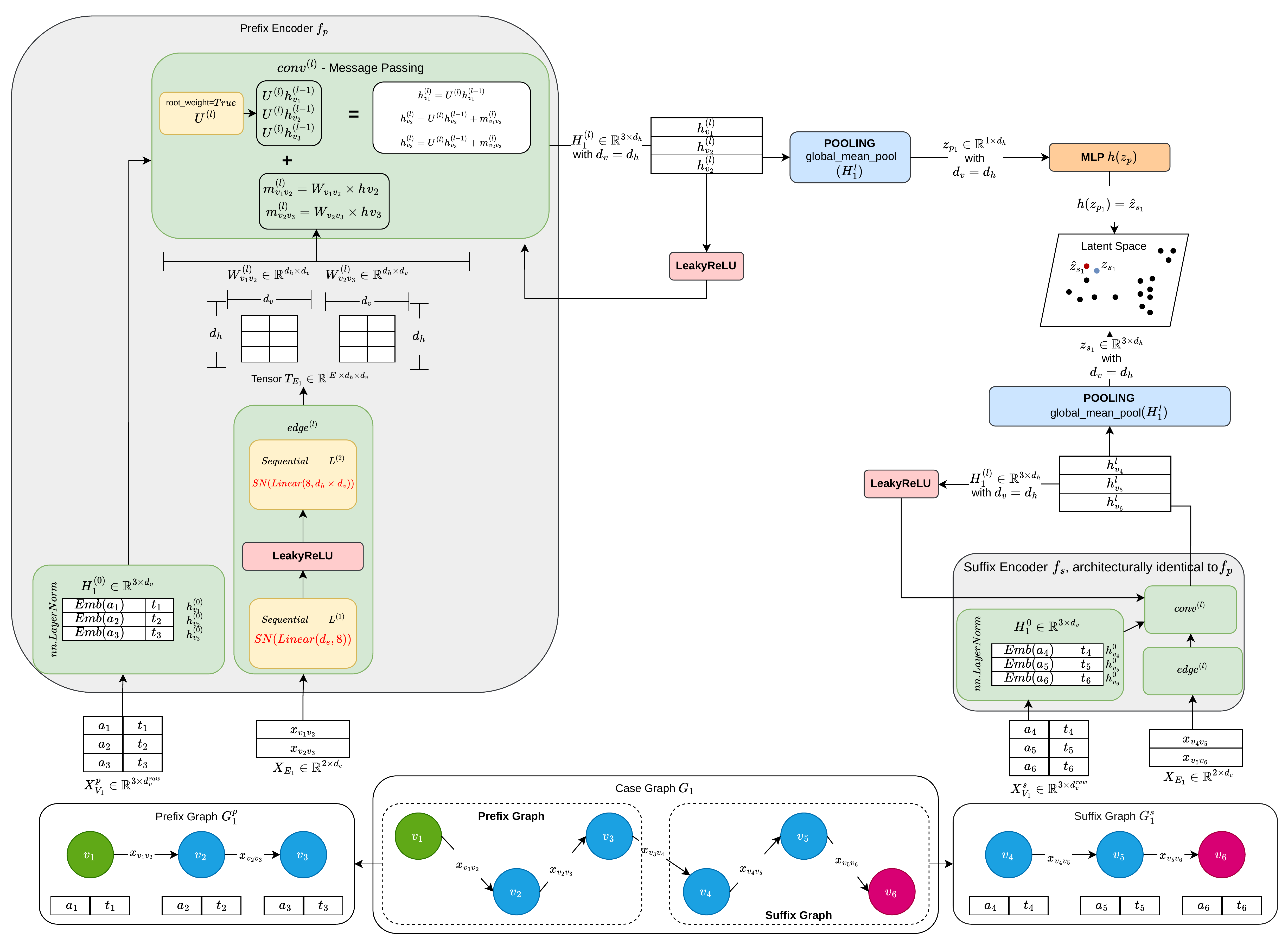}
    \caption{Architecture of the proposed prefix-to-suffix latent retrieval framework. 
A complete case graph \(\smash{G_1}\) is decomposed into a prefix graph \(\smash{G_1^p}\) and a suffix graph \(\smash{G_1^s}\). 
Green nodes denote start events, red nodes denote end events, and blue nodes denote intermediate events. 
Node features combine activity embeddings and temporal descriptors, while edge features encode normalised inter-event durations. 
For each graph, the node-feature matrix \(\smash{H_1^{(0)}}\) is constructed from the initial node attributes and processed by a two-layer edge-conditioned graph encoder. 
The edge network maps each edge attribute to a convolutional filter, which is used in message passing together with the root transformation. 
After non-linear activation and global mean pooling, the prefix and suffix graphs are mapped to graph-level embeddings \(\smash{z_{p_1}}\) and \(\smash{z_{s_1}}\). 
A predictor \(\smash{h(\cdot)}\) then projects the prefix embedding into the suffix latent space, yielding \(\smash{\hat z_{s_1}}\), which is trained to align with the embedding of the true suffix. 
For readability, only one message-passing layer is expanded in detail.}
    \label{fig:architecture}
\end{figure*}

The role of \(h\) is not to decode a suffix explicitly, but to project the prefix representation toward the region of the latent space occupied by plausible future suffixes. The intermediate non-linearity increases the expressive power of this projection, allowing the model to capture non-linear relations between observed prefixes and future continuations. Spectral normalisation is applied to both linear layers in order to control the Lipschitz constant of the predictor and stabilise latent-space transformations. As a result, small perturbations in the prefix embedding are less likely to induce abrupt changes in the predicted suffix embedding \(\hat{z}_s\), which is consistent with the overall objective of learning a smooth and retrieval-friendly latent space.
Figure~\ref{fig:architecture} summarises the proposed framework and details the flow of information from prefix and suffix process graphs to their graph-level latent representations.

\subsection{Training Objective and Adaptive Optimisation}

The objective should be regarded as non-convex due to the combined use of learnable embeddings, edge-conditioned graph convolutions, nonlinear activations, and latent projection layers. Hence, no guarantee of global convexity or convergence to a global optimum can be claimed. Nevertheless, the optimisation remains well behaved in practice: the reconstruction term is smooth in the latent space, the contrastive term uses a soft hinge objective, and spectral normalisation controls the Lipschitz behaviour of the main learned linear mappings.

The model is trained so that the predicted suffix embedding \(\smash{\hat{z}_s}\) matches the true suffix embedding \(z_s\) while remaining separated from a negative suffix embedding \(z_s^{-}\). All three embeddings are \(\ell_2\)-normalised before loss computation: $\smash{\bar{z}=z/{\|z\|_2}}$,
so that optimisation takes place on the unit sphere. In this setting, squared Euclidean distance is a bounded monotone transform of cosine dissimilarity $\smash{\|u-v\|_2^2 = 2 - 2\cos(u,v)}$ with $\smash{\|u\|_2=\|v\|_2=1}$, which makes the objective depend on embedding direction rather than magnitude and is therefore consistent with retrieval.

\subsubsection*{\textbf{Rarity-Aware Reconstruction Loss}}

To preserve semantic alignment between predicted and true suffix representations, we minimize the weighted reconstruction loss
\begin{equation}
\mathcal{L}_{\mathrm{rec}}
=
\frac{\sum_{i=1}^{B} w_i \left\| \bar{\hat{z}}_{s_i} - \mathrm{sg}(\bar{z}_{s_i}) \right\|_2^2}
{\sum_{i=1}^{B} w_i}
\label{eq:lrec}
\end{equation}
where \(B\) is the batch size and \(\mathrm{sg}(\cdot)\) denotes stop-gradient, so that the reconstruction term updates the prefix encoder and predictor while treating the suffix embedding as a fixed target.

To reduce the dominance of frequent behavioural patterns, each sample is weighted according to the rarity of the activities appearing in its prefix and suffix. Let \(\mathcal A_{P_i}\) and \(\mathcal A_{S_i}\) denote the activity sets of the prefix and suffix of sample \(i\). We define \(w_i = \min\! \big(\alpha \max_{a \in \mathcal A_{P_i}} \omega_{\mathrm{pref}}(a) + (1-\alpha)\max_{a \in \mathcal A_{S_i}} \omega_{\mathrm{suf}}(a),\, w_{\max}\big)\), where \(w_{\max}\) is a clipping threshold and \(\alpha \in [0,1]\) controls the relative contribution of prefix and suffix rarity. Since the task focuses on future suffix prediction, suffix rarity is given higher importance. Using the maximum rather than an average preserves the influence of rare activities even in long prefixes or suffixes.

Rarity weights are defined as \(\omega(a)=f(a)^{-\gamma}\), with \(\gamma \in [0,1]\), where \(f(a)\) is the training frequency of activity \(a\). This yields an effective contribution \(f(a)\omega(a)=f(a)^{1-\gamma}\), meaning that \(\gamma\) progressively compresses the dominance of frequent activities without fully erasing the empirical data distribution.

\subsubsection*{\textbf{Soft Contrastive Ranking Loss}}
Let
$d_i^{+}=\left\|\bar{\hat{z}}_{s_i}-\bar{z}_{s_i}\right\|_2^2$ 
and 
$d_i^{-}=\left\|\bar{\hat{z}}_{s_i}-\bar{z}_{s_i}^{-}\right\|_2^2$ 
denote the positive and negative squared distances, respectively. Since the embeddings are normalised, these distances are bounded and behave similarly to cosine-based dissimilarities.
We enforce that the predicted suffix embedding should be closer to the true suffix than to the negative suffix by a margin \(m\). Instead of using a hard hinge, we adopt a temperature-controlled soft hinge:
\begin{equation}
\mathcal{L}_{\mathrm{ctr}}
=
\frac{1}{B}
\sum_{i=1}^{B}
\tau \,
\mathrm{softplus}\!\left(
\frac{d_i^{+}-d_i^{-}+m}{\tau}
\right)
\label{eq:lctr}
\end{equation}
where \(\tau>0\) is a temperature parameter and \(\mathrm{softplus}(x)=\log(1+e^x)\). This formulation preserves non-zero gradients near the decision boundary, unlike a hard hinge whose gradient vanishes immediately once the margin is satisfied.

\subsubsection*{\textbf{Adaptive Margin Estimation}}

Rather than fixing the contrastive margin manually, we estimate it from the empirical distribution of separation gaps $g_i = d_i^{-} - d_i^{+}$.
At the beginning of training, the initial margin is estimated on the training loader from the empirical \(q_0\)-quantile of the gap distribution:
$m^{(0)} = \max\!\left(\mathrm{Quantile}_{q_0}(\{g_i\}),\, 0.1\right)$, with $q_0=0.1$.

During training, the margin is updated at the end of each epoch from the empirical distribution of separation gaps observed over the current epoch:
$m^{(t)} =
\mathrm{clip}\!\left(
\mathrm{Quantile}_{q_t}(\{g_i\}),
\, m_{\min},\,
m_{\max}
\right)$ 
where \(t \in \mathbb{N}\) denotes the training epoch index. The target quantile follows 
$q_t = \min(q_0 + \Delta q \, t,\; q_{\max}).$
This quantile-based schedule gradually increases the separation requirement while keeping the margin grounded in the observed latent geometry. Starting from a low quantile makes the contrastive constraint permissive at the beginning of training, whereas the progressive increase encourages stronger separation later on. The upper cap \(q_{\max}\) avoids over-constraining the model from already well-separated samples and was retained empirically as a stable compromise between latent discrimination and optimisation robustness.

\subsubsection*{\textbf{Gradient-Ratio-Based Loss Balancing}}

The total loss is a weighted combination of reconstruction and contrastive terms:
\begin{equation}
\mathcal{L}
=
\lambda_{\mathrm{rec}} \mathcal{L}_{\mathrm{rec}}
+
\lambda_{\mathrm{ctr}} \mathcal{L}_{\mathrm{ctr}}
\label{eq:ltotal}
\end{equation}
with \(\lambda_{\mathrm{rec}}=1\). Instead of fixing \(\lambda_{\mathrm{ctr}}\), we adapt it dynamically so that the contrastive term remains controlled relative to the reconstruction term on the parameters shared by prefix encoding and prediction: 
$\Theta_{\mathrm{sh}} = \Theta_{f_p} \cup \Theta_{h}$.
We define the corresponding gradient norms 
$G_{\mathrm{rec}}
=
\left\|
\nabla_{\Theta_{\mathrm{sh}}}\mathcal{L}_{\mathrm{rec}}
\right\|_2$, 
$G_{\mathrm{ctr}}
=
\left\|
\nabla_{\Theta_{\mathrm{sh}}}\mathcal{L}_{\mathrm{ctr}}
\right\|_2$.
We define the instantaneous target weight as
$\lambda_{\mathrm{ctr}}^{\star}
=
r_{\max}
{G_{\mathrm{rec}}}/{(G_{\mathrm{ctr}}+\varepsilon)}$, 
where \(r_{\max}\) denotes the target upper ratio between the effective contrastive gradient and the reconstruction gradient, and \(\varepsilon\) is a small numerical constant used for stability. This definition ensures that the contrastive term remains controlled relative to the reconstruction objective, rather than dominating it.
To avoid batch-level oscillations, this instantaneous value is smoothed by exponential moving average: 
$\lambda_{\mathrm{ctr}}^{(t)}
=
\alpha_{\mathrm{ema}}\lambda_{\mathrm{ctr}}^{(t-1)}
+
(1-\alpha_{\mathrm{ema}})\lambda_{\mathrm{ctr}}^{\star}$
where \(\smash{\alpha_{\mathrm{ema}} \in [0,1)}\) is the smoothing coefficient. Larger values of \(\alpha_{\mathrm{ema}}\) increase inertia and yield a more stable but slower adaptation of \(\lambda_{\mathrm{ctr}}\), whereas smaller values make the weighting more reactive to batch-level fluctuations. The resulting coefficient is finally clipped to a maximum value for additional optimisation stability.

\subsubsection*{\textbf{Training Diagnostics}}

In addition to the optimisation objective, we monitor several diagnostic quantities to assess latent-space organisation, optimisation balance, and training stability. These quantities are not additional loss terms, but monitoring signals used to interpret the dynamics of the model during training and validation.

\paragraph{Geometric separation diagnostics}
We first track the average positive and negative squared distances
$\smash{\bar d^{+}}=\smash{\frac{1}{B}\sum_{i=1}^{B} d_i^{+}}$
$\smash{\bar d^{-}}=\smash{\frac{1}{B}\sum_{i=1}^{B} d_i^{-}}$
together with the contrastive retrieval accuracy
$\mathrm{Acc}_{\mathrm{ctr}}
=
\frac{1}{B}\sum_{i=1}^{B}\mathbf{1}(d_i^{+}<d_i^{-})$
which measures how often the true suffix remains closer to the predicted suffix than the negative one.

We also monitor the mean separation gap
$\smash{\bar g = \frac{1}{B}\sum_{i=1}^{B} g_i}$, 
$\smash{g_i=d_i^{-}-d_i^{+}}$
as well as the fraction of active contrastive constraints,
$\smash{r_{\mathrm{active}}
=
\frac{1}{B}\sum_{i=1}^{B}
\mathbf{1}_{\{d_i^{+}-d_i^{-}+m>0\}}}$
which indicates the proportion of samples for which the margin constraint is still effectively active. To detect weak separation or latent collapse, we further track the empirical dispersion of the gap distribution through its standard deviation and selected quantiles (e.g., \(\smash{p_{10}}\), \(\smash{p_{50}}\), \(\smash{p_{90}}\)).

\paragraph{Gradient interaction and balancing diagnostics}
To evaluate whether reconstruction and contrastive objectives are cooperative or conflicting on the shared parameters, we compute the cosine similarity between their gradient vectors:
\begin{equation}
\cos(\mathcal{L}_{\mathrm{rec}},\mathcal{L}_{\mathrm{ctr}})
=
\frac{
\left\langle
\nabla_{\Theta_{\mathrm{sh}}}\mathcal{L}_{\mathrm{rec}},
\nabla_{\Theta_{\mathrm{sh}}}\mathcal{L}_{\mathrm{ctr}}
\right\rangle
}{
\left\|
\nabla_{\Theta_{\mathrm{sh}}}\mathcal{L}_{\mathrm{rec}}
\right\|_2
\left\|
\nabla_{\Theta_{\mathrm{sh}}}\mathcal{L}_{\mathrm{ctr}}
\right\|_2
+\varepsilon
}
\label{eq:grad_cos}
\end{equation}
Positive values indicate cooperative directions, whereas negative values reveal gradient interference.

We additionally monitor the corresponding gradient norms $\smash{G_{\mathrm{rec}}}$ and $\smash{G_{\mathrm{ctr}}}$, their raw ratio
$\smash{\rho_{\mathrm{raw}}={G_{\mathrm{ctr}}}/({G_{\mathrm{rec}}+\varepsilon})}$ %
and the effective ratio after adaptive weighting,
$\smash{\rho_{\mathrm{eff}}}=\smash{
{\lambda_{\mathrm{ctr}}G_{\mathrm{ctr}}}/
{(\lambda_{\mathrm{rec}}G_{\mathrm{rec}}+\varepsilon)}}$
which indicates whether the contrastive objective remains controlled relative to the reconstruction objective during training.

\subsection{Inference and Recommendation}

At inference time, a prefix graph \(\smash{G^p}\) is first encoded into a prefix embedding and projected into the suffix latent space:
$\smash{z_p = f_p(G^p)}$ and 
$\smash{\hat{z}_s = h(z_p)}$. 
To remain consistent with training, the predicted embedding and the candidate suffix embeddings are \(\smash{\ell_2}\)-normalised before retrieval.

Let 
$\smash{\mathcal{C}=\left\{\gamma_1,\ldots,\gamma_K\right\}}$ 
denote the set of candidate suffixes, and let \(\smash{z_{s_j}}\) be the latent embedding associated with candidate suffix \(\smash{\gamma_j}\). Complete suffix prediction is then performed through nearest-neighbour retrieval in the latent suffix space:
\begin{equation}
\mathcal{N}_k(\hat{z}_s)
=
\operatorname{TopK}_{\gamma_j \in \mathcal{C}}
\left(
-\|\hat{z}_s - z_{s_j}\|_2^2
\right)
\label{eq:topk}
\end{equation}
where \(\smash{\mathcal{N}_k(\hat{z}_s)\subseteq \mathcal{C}}\) denotes the set of the top-\(k\) retrieved suffixes. Since all embeddings are normalised, this ranking is equivalent to cosine-based retrieval up to a monotone transformation.

The top-\(1\) retrieved suffix is used for direct complete suffix prediction: 
$\smash{\hat{\gamma}^{(m)}
=
\argmin_{\gamma_j \in \mathcal{C}}
\|\hat{z}_s - z_{s_j}\|_2^2}$.

Beyond direct prediction, the retrieved top-\(k\) set can also support recommendation. In particular, given a target KPI scoring function \(q(\gamma_j)\), the final recommended suffix can be selected among the retrieved plausible futures as
\begin{equation}
\gamma^{\star}_{\mathrm{KPI}}
=
\argmin_{\gamma_j \in \mathcal{N}_k(\hat{z}_s)} q(\gamma_j)
\label{eq:kpi_inference}
\end{equation}
for example when optimising total trace duration. This formulation allows the model to distinguish between prediction, which returns the nearest latent suffix, and recommendation, which selects among plausible retrieved continuations with respect to an operational objective.

\section{Experiments}

\subsection{Experimental Settings}

\paragraph{Datasets}
We evaluate the framework on two real-world event logs: BPIC2019\footnote{\url{https://data.4tu.nl/articles/_/12715853/1}} (large-scale purchase order process), and BPIC2017\footnote{\url{https://data.4tu.nl/articles/dataset/BPI_Challenge_2017/12696884}} (loan application process using the standard activity-level abstraction). To ensure valid suffix retrieval, we first filter out incomplete or non-compliant cases using business compliance rules. For a consistent evaluation, we discard dataset-specific attributes and retain only the case identifier, activity label, and event timestamp. From these, we extract node-level cyclical calendar encodings and edge-level inter-event durations, transforming each trace into prefix--suffix graph pairs. The main statistics are reported in Table~\ref{tab:dataset_stats}.

\begin{table}[!ht]
\caption{Main statistics of the evaluated datasets}
\label{tab:dataset_stats}
\centering
\scriptsize
\begin{tabular}{lcc}
\toprule
Statistics & BPIC2019 & BPIC2017 \\
\midrule
\#Cases & 251,734 & 31,509 \\
\#Events & 1,595,923 & 1,202,267 \\
\#Activities & 39 & 25 \\
\#Variants & 11909 & 4047 \\
Avg. Case Len. & 6.34 & 17.7 \\
\bottomrule
\end{tabular}
\end{table}

\paragraph{Evaluation metrics}
We evaluate the model from five complementary perspectives: ranking ($\mathrm{Recall@}1$, $\mathrm{Recall@5}$, $\mathrm{MRR@}5$), semantic (normalised DLD), temporal (MAE), and KPI-aware recommendation qualities.

For ranking quality, let \(r_i\) denote the rank of the reference suffix for query \(i\). Since each query is associated with a single ground-truth suffix, $\mathrm{Recall@}k$ is equivalent to Hit Rate ($\mathrm{HR@}k$) and is defined as
$\mathrm{Recall@}k = \mathrm{HR@}k
=
\frac{1}{N}\sum_{i=1}^{N}\mathbf{1}(r_i \le k)$ 
and Mean Reciprocal Rank at cutoff \(k\), defined as
$\mathrm{MRR@}k
=
\frac{1}{N}\sum_{i=1}^{N}
\frac{\mathbf{1}(r_i \le k)}{r_i}$ 
where \(N\) is the number of evaluated prefixes. In contrast, $\mathrm{NDCG@}k$ additionally accounts for the rank position of the reference suffix within the top-\(k\) retrieved candidates, assigning higher scores when it appears closer to the top of the ranking.

For recommendation-oriented evaluation, we additionally consider a KPI-aware selection protocol. Given the top-\(k\) retrieved plausible suffixes for a prefix, the final recommended suffix is selected according to a target KPI \(q(\gamma)\). We then report recommendation-specific metrics such as the average gain with respect to the real suffix duration, the percentage of improved cases, the percentage of cases for which a recommendation is found, and the compliance rate of the selected recommendations.

\paragraph{Evaluation protocol}
We report two complementary retrieval settings. In \emph{local sampled retrieval}, the true suffix is ranked against a sampled set of negatives (e.g., \(1\) positive vs. \(199\) negatives). In \emph{global retrieval}, the true suffix is ranked against the full suffix repository, which may contain several hundred thousand candidates.
In our experiments, the KPI is the total trace duration, so that recommendation amounts to selecting, among the retrieved plausible futures, the suffix with minimal predicted duration. Unless otherwise stated, we use \(k=15\) for this KPI-aware recommendation analysis.

\begin{table}[!ht]
\caption{Main structural suffix prediction results. Lower N-DLD and MAE indicate better performance. Case-Level Retrieval.}
\label{tab:main_results_2datasets}
\centering
\scriptsize
\setlength{\tabcolsep}{5pt} 
\begin{tabular}{lccccc}
\toprule
Method & R@1 & R@5 & MRR@5 & N-DLD$^{\dagger}$ & MAE$^{\ddagger}$ \\
\midrule
\multicolumn{6}{c}{\textbf{BPIC2019 (Purchase Order Process)}} \\
\midrule
SEP-LSTM    & - & - & - & 0.16 & 509.5 \\
ED-LSTM     & - & - & - & 0.15 & 516.6 \\
CRTP-LSTM (NDA)$^{\P}$   & - & - & - & 0.15 & 488.7 \\
SuTraN (NDA)$^{\P}$     & - & - & - & 0.13 & 486.8 \\
NBA         & - & - & - & 0.43 & - \\
\textbf{Proposed method} & \textbf{0.55} & \textbf{0.89} & \textbf{0.70} & \textbf{0.09} & \textbf{427.8} \\
\midrule
\multicolumn{6}{c}{\textbf{BPIC2017 (Loan Application)}} \\
\midrule
SEP-LSTM    & - & - & - & 0.78 & 197.1 \\
ED-LSTM     & - & - & - & 0.68 & 202.6 \\
CRTP-LSTM (NDA)$^{\P}$   & - & - & - & 0.59 & 148.4 \\
SuTraN (NDA)$^{\P}$     & - & - & - & 0.62 & \textbf{147.6} \\
\textbf{Proposed method} & \textbf{0.47} & \textbf{0.82} & \textbf{0.62} & \textbf{0.53} & 192.8 \\
\bottomrule
\end{tabular}

\vspace{1mm}

\raggedright

\footnotesize{$^{\dagger}$ When a baseline reports normalised Damerau--Levenshtein similarity (N-DLS) instead of N-DLD, we convert it as \(\mathrm{N\text{-}DLD}=1-\mathrm{N\text{-}DLS}\). \\

$^{\ddagger}$ All MAE Remaining Runtime Prediction values are reported in hours. When originally given in seconds or minutes, they are converted accordingly for consistency.
\\

$^{\P}$ NDA stands for non-data-aware.}

\end{table}

\paragraph{Implementation details}
Experiments employ two-layer GNN encoders (\texttt{NNConv}, global mean pooling) and a two-layer MLP predictor with hidden width \(8\) mapping to a 128-dimensional shared latent space ($d_z=d_h=128$). Spectral normalisation is applied to both the predictor and edge-filter networks to stabilise latent geometry. Models are trained on an 80/10/10 dataset split using Adam (learning rate $10^{-3}$, batch size 32) with early stopping. Embeddings are \(\ell_2\)-normalised before computing the reconstruction and contrastive losses. Rarity-aware weighting uses \(\omega(a)=f(a)^{-\gamma}\), with prefix--suffix mixing coefficient \(\alpha=0.2\) and clipping threshold \(w_{\max}=3.0\). The contrastive objective uses a soft hinge with temperature \(\tau=0.1\), an adaptive margin clipped to \([m_{\min},m_{\max}]=[0.1,1.0]\), and a progressive quantile schedule with \(q_0=0.1\), \(\Delta q=0.02\), and \(q_{\max}=0.3\). Loss balancing is controlled dynamically through gradient norms with target ratio \(r_{\max}=0.5\) and exponential moving average coefficient \(\alpha_{\mathrm{ema}}=0.95\). A small numerical constant \(\varepsilon=10^{-12}\) is used in gradient-ratio and gradient-cosine computations for numerical stability.

\paragraph{Baselines}
We compare the proposed framework against two groups of baselines.

\textbf{Structural suffix prediction baselines:}
These baselines evaluate semantic accuracy, temporal plausibility, and, when applicable, retrieval quality. They cover the main families of prior work: \emph{SEP-LSTM} \cite{Tax_2017} for iterative next-step rollout, \emph{ED-LSTM} \cite{taymouri2021deepadversarialmodelsuffix} for autoregressive suffix generation, \emph{CRTP-LSTM} \cite{articleGunnarsson} for direct complete suffix prediction, \emph{SuTraN} \cite{inproceedingsWuyts} for transformer-based suffix and remaining-time prediction, and \emph{NBA} \cite{Weinzierl_2020} for retrieval-related prescriptive monitoring baseline.

\textbf{KPI-aware recommendation baselines:}
These baselines evaluate recommendation quality under the operational objective of throughput-time minimisation. Since no directly comparable complete-suffix retrieval baselines are available for this setting, we compare against prescriptive approaches explicitly targeting throughput time that provide the closest available baselines for KPI-aware process recommendation: \emph{NBA} \cite{Weinzierl_2020}, which uses nearest-neighbour-based candidate selection, and \emph{Goal-oriented NBA} \cite{agarwal2022goalorientedbestactivityrecommendation}, which adds goal satisfaction and conformance constraints.

\begin{table}[!ht]
\caption{KPI-aware recommendation. KPI = trace duration.}
\label{tab:kpi_reco_comparison}
\centering
\footnotesize
\setlength{\tabcolsep}{6pt}
\begin{tabular}{lccc}
\toprule
Method & DLD$^{\S}$ & \% Improved$^{\star}$ & \% Compliant \\
\midrule
\multicolumn{4}{c}{\textbf{BPIC2019 (Purchase Order Process)}} \\
\midrule
NBA               & $\approx 0.43$ & 19.88 & 50.74 \\
Goal-oriented NBA & $\approx 0.11$ & \textbf{86.48} & 100.00 \\
Proposed method   & \textbf{0.09}  & 71.91          & \textbf{100.00} \\
\bottomrule
\end{tabular}

\vspace{1mm}
\raggedright
\tiny{$^{\star}$ The reported \% Improved corresponds to the published \(GS_{\mathrm{pred}}\%\), i.e., the percentage of recommended sequences satisfying the throughput-time objective. This metric is used here as a KPI-oriented proxy and should not be interpreted as a strict improvement rate with respect to the factual continuation. For these methods, \% Compliant corresponds to the reported conformance rate \(C\%\). \\
$^{\S}$ DLD values are approximate mean visual estimates for both the NBA of Weinzierl et al. \cite{10.1007/978-3-030-94343-1_3} and the method of Agarwal et al. \cite{agarwal2022goalorientedbestactivityrecommendation}, derived from the DL-distance curves reported in \cite{agarwal2022goalorientedbestactivityrecommendation}, since no exact aggregate values are tabulated in the paper. The plotted values lie in \([0,1]\), which suggests a normalised Damerau--Levenshtein distance.}
\end{table}

\subsection{Results}
\paragraph{Performance Comparison}

Table~\ref{tab:main_results_2datasets} reports the main comparison across the two evaluated datasets. First, the retrieval-oriented ranking metrics provide direct evidence that the learned latent space is informative. On BPIC2019, the proposed framework achieves Recall@1 = 0.55, Recall@5 = 0.89, and MRR@5 = 0.70, while on BPIC2017 it reaches Recall@1 = 0.47, Recall@5 = 0.82, and MRR@5 = 0.62. Since these scores are computed for retrieval of the \emph{exact suffix}, they indicate that the true continuation is not only frequently retrieved, but also typically ranked very high among the candidate suffixes on both datasets. In particular, the MRR values show that, when the exact suffix is retrieved, it tends to appear close to the top of the ranking rather than in lower positions. Hence, these metrics support the quality of the latent retrieval space, although they are not directly comparable to the generation-based baselines reported in the literature.

The comparison with prior work should instead be interpreted through the shared structural and temporal metrics. On BPIC2019, the proposed method achieves the best semantic result. This shows that latent retrieval preserves the structural identity of future continuations more accurately than both sequential generation and existing retrieval-oriented alternatives. On BPIC2017, the same trend holds: the proposed framework again obtains the lowest semantic error.

The temporal comparison is more nuanced. On BPIC2019, the proposed method also outperforms the baselines in MAE, showing that the retrieval framework can recover suffixes that are not only structurally accurate but also temporally well aligned with the historical continuation. On BPIC2017, however, the temporal comparison is less favourable. This difference is consistent with the objective of the framework. Unlike sequence-generation approaches that are optimised to reproduce the realised temporal trajectory as closely as possible, our method retrieves \emph{historically plausible} suffixes from a candidate space. As a consequence, a retrieved suffix may remain structurally accurate and operationally valid while still differing from the factual suffix in total duration. When temporal error is higher, it should therefore not be interpreted solely as a weakness, but rather as the consequence of a recommendation-oriented setting in which the goal is to surface plausible alternatives instead of reproducing one observed future exactly. This interpretation is consistent with Table~\ref{tab:kpi_reco}, where the retrieved candidates prove useful for recommendation under a duration-oriented objective. Interestingly, some retrieved recommendations correspond to unseen but compliant variants. When they remain valid under the process constraints and lead to better KPI values, they can be viewed as plausible process improvements rather than simple memorised continuations.

Table~\ref{tab:kpi_reco_comparison} provides an additional positioning with respect to prescriptive recommendation baselines from the literature. On BPIC2019, the proposed framework achieves the best structural quality, as reflected by the lowest DLD, while remaining fully compliant. The comparison on \% Improved should however be interpreted more cautiously, since the published NBA-based scores are KPI-oriented success rates that are not perfectly aligned with our recommendation-found criterion. As a result, this table mainly shows that the proposed method offers a favourable trade-off between suffix fidelity and compliance in KPI-aware recommendation.

\begin{table}[!ht]
\caption{KPI-aware recommendation ($k=15$ retrieval, trace duration, in hours).}
\label{tab:kpi_reco}
\centering
\small
\begin{tabular}{lc}
\toprule
Metric & Value \\
\midrule
Number of evaluated cases (test set) & 68648 \\
Average real duration & 1111.4 \\
Average recommended duration & 437.3 \\
Average gain & 674.1 \\
Median gain & 502.5 \\
Pct. recommendation found & 71.91\% \\
Pct. recommended compliant & 100\% \\
Pct. seen variants & 72.98\% \\
Pct. unseen variants & 27.02\% \\
\bottomrule
\end{tabular}
\end{table}

\begin{figure*}[t]
    \centering
    \begin{minipage}[t]{0.49\textwidth}
        \centering
        \includegraphics[width=\textwidth, trim={0cm 0cm 0cm 1.5cm}, clip]{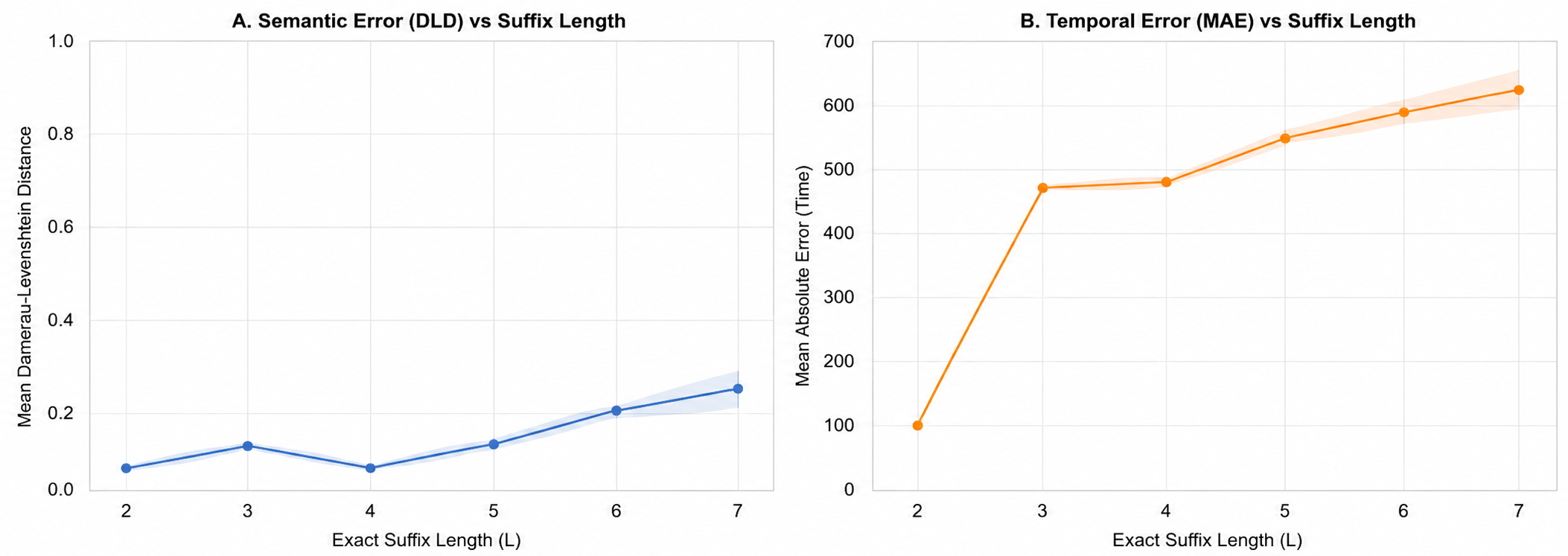}
 %
    \end{minipage}
    \hfill
    \begin{minipage}[t]{0.49\textwidth}
        \centering
        \includegraphics[width=\textwidth, trim={0cm 0cm 0cm 0.64cm}, clip]{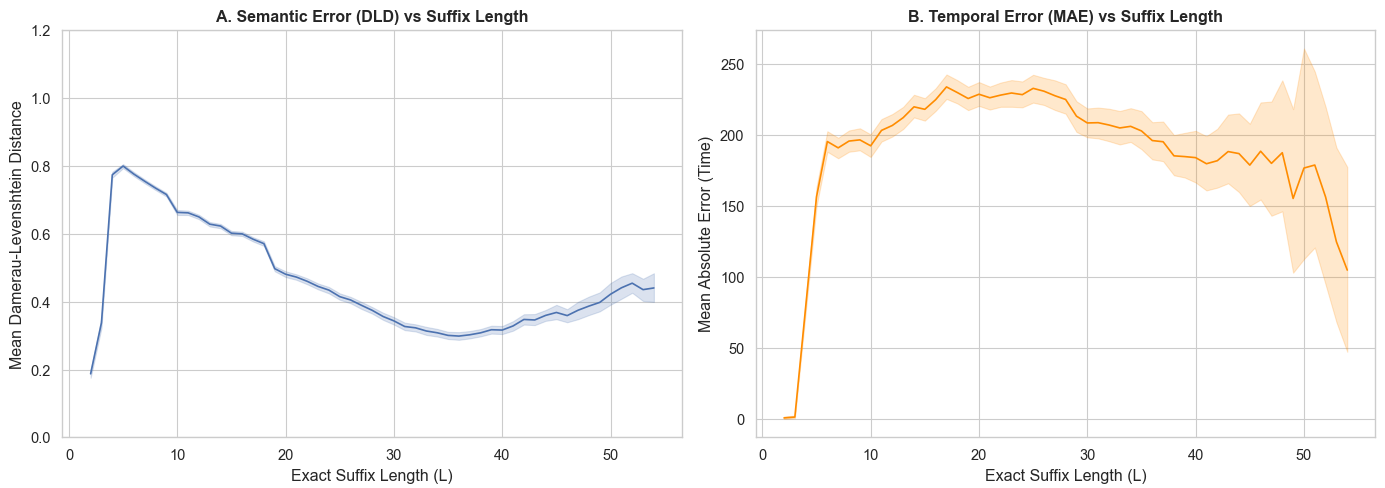}
    \end{minipage}
    \caption{Disaggregated semantic and temporal performance of the proposed framework. (Left two plots) BPIC2019. (Right two plots) BPIC2017. (Blue plots) Mean semantic error, measured by normalised Damerau--Levenshtein distance (DLD), as a function of exact suffix length. (Orange plots) Mean temporal error (MAE) by suffix length. On each plot, shaded area display the 95\% level confidence intervals.}
    \label{fig:length_rarity_analysis}
\end{figure*}

\paragraph{Robustness Across Suffix Length and Variant Rarity}
Figure~\ref{fig:length_rarity_analysis} evaluates the framework's robustness across suffix length and variant rarity. As suffix length increases, semantic error grows gradually, indicating that the latent retrieval mechanism successfully preserves neighbourhood structures over longer horizons. Yet, temporal error (MAE) degrades more sharply on longer continuations due to high duration variance in BPIC2019, challenging even the variant-level median oracle. Similarly, across rarity groups, performance decreases progressively from head to tail variants, confirming that the latent space remains informative even for rare behaviours.

Table~\ref{tab:standard_complex} highlights performance on standard versus complex cases. While a frequency-based baseline excels on standard cases through memorisation, it collapses on complex cases. In contrast, the proposed framework maintains robust predictive ability in these difficult, non-routine scenarios.

\begin{table}[t]
\caption{Comparison with a frequency-based baseline.}
\label{tab:standard_complex}
\centering
\scriptsize
\begin{tabular}{lcc}
\toprule
Group & Baseline Recall@1 & Proposed Recall@1 \\
\midrule
Standard cases & 1 & 0.86 \\
Complex cases  & 0.0 & 0.56 \\
\bottomrule
\end{tabular}
\end{table}

\subsection{Sensitivity and Stability Analysis}
Although the final configuration was selected through validation-guided experimentation and the main experiments were repeated across multiple runs to verify the consistency of the observed trends, we do not report the complete hyperparameter sensitivity analysis. Extending the empirical study in that direction would provide a more exhaustive characterisation of the robustness of the proposed framework.

Each reported result corresponds to 10 independent training runs with different random seeds. Experiments were conducted on Intel(R) Core(TM) i7-10750H CPU at 2.60 GHz, with 8 GB RAM, using a single NVIDIA GeForce RTX 2070 8 GB GPU. The contrastive negative construction over the suffix repository was accelerated through triangle inequality and GPU.

\paragraph*{\textbf{Complexity considerations}}
Let \(M\) denote the number of prefix-suffix training pairs, \(K\) the candidate-pool size used for contrastive negative selection, \(|V|\) and \(|E|\) the numbers of nodes and edges in a graph, and \(d_h\) the hidden dimension. A single encoder forward pass is linear in the graph size, i.e., \(O(|E|d_h + |V|d_h)\) up to constant factors induced by the edge networks. Training therefore scales linearly with the number of processed prefix and suffix graphs. The main additional cost comes from contrastive negative construction, whose offline ranking stage scales with the size of the candidate repository and was implemented with hardware-accelerated batched computation. In memory, graph storage scales linearly with the total numbers of nodes and edges, while latent repositories scale as \(O(|\mathcal{C}|d_z)\). The effective training sample size is the number of generated prefix-suffix pairs after filtering, i.e. 688\,396 for BPIC2019 and 725\,952 for BPIC2017.

\subsection{Ablation Study}
The ablation study is central to the methodological claim of the paper because the proposed retrieval space relies on two design choices: process-aware hard negatives and Lipschitz-oriented spectral normalisation. We thus evaluate two controlled variants on BPIC2019: (i) removing the process-aware hard negative strategy, and (ii) removing spectral normalisation from the edge networks and prefix-to-suffix predictor.

The ablation \emph{w/o process-aware hard negatives} removes only the difficulty-aware selection strategy, sampling negatives from the most distant (and easily separable) suffixes. In contrast, the full model samples negatives from a controlled rank band of the candidate pool. This yields harder, behaviourally plausible contrastive examples, forcing the model to discriminate against structurally close alternatives rather than just clearly unrelated futures.

The second ablation, \emph{w/o Lipschitz regularisation}, removes spectral normalisation from the edge networks and the prefix-to-suffix predictor while preserving the underlying architecture. This variant isolates the effect of spectral normalisation, evaluating how constraining the sensitivity of edge-conditioned filters and latent projections impacts representation stability and retrieval quality.

\begin{table}[!ht]
\caption{Ablation Study of the framework. Variant-Level Retrieval.}
\label{tab:ablation}
\centering
\scriptsize
\setlength{\tabcolsep}{3.5pt} 
\begin{tabular}{lccccc}
\toprule
Variant & Recall@1 & Recall@5 & MRR@5 & N DLD$_{\text{Top-1}}$ & MAE \\
\midrule
w/o Hard Neg.   & 0.08 & 0.31 & 0.20 & 0.29 & 954 \\
w/o Lipschitz   & 0.43 & 0.83 & 0.58 & 0.23 & 827 \\
Full model      & \textbf{0.83} & \textbf{0.97} & \textbf{0.89} & \textbf{0.09} & \textbf{428} \\
\bottomrule
\end{tabular}
\end{table}

Table~\ref{tab:ablation} reports the corresponding results. Overall, both ablations lead to a degradation of performance, confirming that the full model benefits from both components. Removing process-aware hard negatives weakens the discriminative structure of the latent suffix space, which suggests that the contrastive signal is substantially more informative when negative suffixes are selected among behaviourally plausible but distinct alternatives rather than sampled in a simpler or less structured manner. In other words, the gain does not only come from contrasting positives against arbitrary negatives, but from contrasting them against suffixes that are sufficiently close to be challenging while still belonging to different process continuations.
The removal of Lipschitz regularisation also degrades performance, with a particularly visible effect on MAE. This suggests that spectral normalisation contributes not only to semantic stability, but also to the temporal coherence of the learned latent space. Since edge attributes directly encode inter-event durations, constraining the sensitivity of the edge-conditioned filters likely helps prevent excessive temporal distortions in the graph representations. This in turn improves the temporal plausibility of the retrieved suffixes.

\section{Conclusion}

This paper proposes a graph-based latent retrieval framework for complete suffix prediction in business processes. This directly addresses several limitations of prior work: instead of autoregressive generation, prefixes and suffixes are modelled as directed attributed process multigraphs and mapped into a shared latent space using edge-conditioned GNNs, stabilised by Lipschitz-oriented spectral normalisation. By projecting prefix embeddings to retrieve the nearest suffix representations, the framework preserves both event- and transition-level information. Furthermore, this retrieval-based formulation inherently supports top-$k$ plausible futures and enables KPI-aware recommendations, such as optimising trace duration. Future work will explore LTL compliance rules, richer context, and causal recommendation settings.


\bibliographystyle{IEEEtran}
\bibliography{biblio_2}

\end{document}